\documentclass[journal]{IEEEtran}

\usepackage{amsmath,bm}
\usepackage{array}
\usepackage{amssymb,amsfonts}
\usepackage{booktabs}
\usepackage[all,arc]{xy}
\usepackage{enumerate}
\usepackage{mathrsfs}
\usepackage[pass]{geometry}
\usepackage{titletoc}
\usepackage[noend]{algpseudocode}
\usepackage{graphicx}
\usepackage[caption=false,font=footnotesize]{subfig}
\usepackage{fixltx2e}
\usepackage{algorithm}
\usepackage{lipsum}
\usepackage{dblfloatfix} 
\usepackage{cite}
\usepackage{enumitem}
\usepackage{booktabs} 
\usepackage{multirow}
\usepackage{xcolor}
\usepackage{balance}

\usepackage{cite}

\ifCLASSINFOpdf
\else
\fi

\begin{document}
%
\title{Few-Shot Bearing Fault Diagnosis Based on Model-Agnostic Meta-Learning}
\author{Shen~Zhang,~\IEEEmembership{Member,~IEEE,}
    Fei~Ye,~\IEEEmembership{Member,~IEEE,}
	Bingnan~Wang,~\IEEEmembership{Senior Member,~IEEE,}
    and~Thomas~G.~Habetler,~\IEEEmembership{Fellow,~IEEE}
\thanks{S. Zhang, and T. G. Habetler are with the School
of Electrical and Computer Engineering, Georgia Institute of Technology, Atlanta, GA 30332, USA (e-mail: shenzhang@gatech.edu, thabetler@ece.gatech.edu).}
\thanks{F. Ye is with California PATH, University of California, Berkeley, CA 94720, USA (e-mail: fye@berkeley.edu).}
\thanks{B. Wang is with the Mitsubishi Electric Research Laboratories, 201 Broadway, Cambridge, MA 02139, (e-mail: bwang@merl.com).}
%
}

%


\maketitle
%
\begin{abstract}
The rapid development of artificial intelligence and deep learning has provided many opportunities to further enhance the safety, stability, and accuracy of industrial Cyber-Physical Systems (CPS). As indispensable components to many mission-critical CPS assets and equipment, mechanical bearings need to be monitored to identify any trace of abnormal conditions. Most of the data-driven approaches applied to bearing fault diagnosis up-to-date are trained using a large amount of fault data collected \textit{a priori}. In many practical applications, however, it can be unsafe and time-consuming to collect sufficient data samples for each fault category, making it challenging to train a robust classifier. In this paper, we propose a few-shot learning framework for bearing fault diagnosis based on model-agnostic meta-learning (MAML), which targets for training an effective fault classifier using limited data. In addition, it can leverage the training data and learn to identify new fault scenarios more efficiently. Case studies on the generalization to new artificial faults show that the proposed framework achieves an overall accuracy up to 25\% higher than a Siamese-network-based benchmark study. Finally, the robustness and the generalization capability of the proposed framework is further validated by applying it to identify real bearing damages using data from artificial damages, which compares favorably against 6 state-of-the-art few-shot learning algorithms using consistent test environments.
\end{abstract}
%
%
%
\IEEEpeerreviewmaketitle
\section{Introduction}
%
%

Cyber-Physical Systems (CPS) are a mixture of computation, networking, and physical processes, in which the embedded computational algorithms and networks have the power to monitor and control the physical components \cite{maml_icems, AICPS}. The rapid development of artificial intelligence (AI) and the Internet of Things provides further opportunities to advance the technology frontier for sensing, monitoring, and interpreting CPS deployed in industry \cite{CPS, CPS1}. As the essential component of many physical systems with rotating equipment, mechanical bearings are widely used in a variety of safety-critical applications such as airplanes, vehicles, production machinery, wind turbines, air-conditioning systems, and elevator hoists \cite{IoT}. Therefore, the health conditions of bearings have major impacts on the reliability and performance of many industrial CPS \cite{CPS1}.

While many data-driven and AI-based technologies have been applied to enhance the accuracy and reliability of bearing fault diagnosis \cite{review1, review2, review3}, most of them require a large amount of training data such as vibration \cite{Paderborn_paper, Siamese, CaAE}, acoustic \cite{AE1, AE2}, and motor current \cite{current1, current2} signals. In real-world applications, however, it is often impossible to obtain sufficient data samples to train a robust classifier that identifies every fault type \cite{Siamese}. One of the reasons is that most bearing degradation would evolve slowly over time, a process that generally takes months or even years. Furthermore, certain safety-critical applications may not be allowed to run into faulty states \cite{Siamese}. Therefore, it can be expensive, unsafe, and often impractical to collect a sufficient amount of data at each bearing fault condition, which will inevitably result in data imbalance issues \cite{RUL, GAN2}. All of these limitations on real-world bearing fault detection require the use of more effective algorithms that can leverage the limited data to train bearing fault classifiers with good generalization capabilities.

To achieve this goal, one approach is to leverage the limited data available at each class to perform data augmentation, using classical methods like signal translation and time stretching \cite{data-translation}, or the more recent generative adversarial network (GAN) \cite{GAN1, GAN2}. The quality and temporal dependency of the generated time-series data is worth further validation \cite{reza2020}. It has been reported in \cite{GAN2} that ``the quality of generated spectrum samples'' from a GAN ``isn't good enough to provide auxiliary information'', which turns out to actually reduce the classifier accuracy after including these  generated data into the training process. Therefore, the effectiveness of this method highly depends on the quality and amount of real data to generate high-quality ``fake'' data. Additionally, the well-known issues related to GAN training, such as instability, mode collapse, and weak gradient \cite{GAN_training} can further impact their performance to monitor the health status of bearings in industrial CPS-based applications. Inspired by the GAN structure, \cite{adversarial_learning} applies adversarial learning to extract generalized features and proposes an instance level weighted mechanism to reflect the similarities of testing samples with known health states. The unknown fault mode can be effectively identified through domain adaptation.

Besides data augmentation techniques, another promising approach to mitigate the limited data issue is to apply few-shot or one-shot learning methods \cite{reza2020}, which has been successfully applied to a wide variety of tasks including few-shot image recognition, path planning of autonomous agents, and more recently in fault diagnosis \cite{Siamese,  Prototypical, CaAE}. Few-shot learning evaluates the model's generalization capability to classes not previously seen in the training process, given only a few samples of each new class \cite{Prototypical}. Therefore, few-shot learning methods are well-suited to tackle the data imbalance issue as we can train a model that generalizes to the imbalanced classes.

Among the existing work that applies few-shot learning methods to bearing fault diagnosis \cite{Siamese,  Prototypical, CaAE, Few_shot_transfer}, \cite{Siamese} proposed a Siamese neural network-based model that demonstrated enhanced fault diagnosis performance when only using 9 training samples per class. Additionally, \cite{Prototypical} applies a deep prototypical network-based method for few-shot bearing fault diagnosis. Despite demonstrating better performance when compared with supervised learning methods, all of the bearing fault types in the test set have shown up in the training set, as they are only differed by their defect diameters and operating conditions. In \cite{CaAE}, an auto-encoder and a capsule network (CaAE) are implemented for the same purpose. However, the case studies in \cite{CaAE} are not formulated in the standard context of few-shot learning, as all of its identified bearing fault classes are already seen in the training process. 

Another promising application is to leverage the generalization capability of few-shot learning to identify real bearing failures using data collected on artificially-damaged bearings, which can be a more convenient and cost-effective way when compared with collecting sufficient data on naturally-evolved failures. Despite its significance of the promising outcome, this Artificial-to-Real problem has only been investigated in a very recent few-shot learning work \cite{Few_shot_transfer} using few-shot transfer learning algorithms such as feature transfer, frozen parameters, and relation nets. While many efforts are made to address this problem using transfer learning to perform domain-adaptation methods including parameter fine-tuning \cite{parameter_tuning}, deep inception net with atrous convolution \cite{ACDIN}, capsule networks \cite{capsule}, adversarial learning \cite{adversarial_learning}, metric-based meta learning \cite{wang2020few, wang2021metric}, and convolutional neural networks aided by conditional data alignment and unsupervised prediction consistency techniques \cite{li2020deep}, there is still room to further improve the model's generalization capability when transferring to realistic failures. For example, it has been reported in \cite{capsule} that when compared to Artificial-to-Artificial tasks that aim at generalizing from the trained artificial faults to new artificial ones, there is an obvious decrease in the average accuracy for Artificial-to-Real tasks, which may have been caused ``by the differences between the artificial damage and the natural damage'' \cite{capsule}.

Therefore, to further mitigate the limited data issue and improve the model's generalization capability, this paper seeks to achieve effective fault diagnosis using the minimum amount of data with model-agnostic meta-learning (MAML) \cite{maml_icems}. Beyond just generalizing to new tasks more effectively, MAML can also learn the process of learning itself, or learning to learn. Specifically, MAML is explicitly designed to train the model’s initial parameters such that ``the model has maximal performance on a new task after the parameters have been updated through one or more gradient steps computed with a small amount of data from that new task'' \cite{MAML}. 

We summarize the main contributions of this paper as follows:
\begin{enumerate}
    \item We propose a few-shot learning framework for bearing fault diagnosis with limited data, which is achieved by developing a diagnostic framework based on MAML.
    \item We demonstrate that the MAML-based few-shot diagnostic framework consistently outperforms state-of-the-art results using the Siamese Network \cite{Siamese} when generalizing to new artificial faults. For example, with 3 unseen classes and using a small training set with only 9 samples per class, our method can achieve an average accuracy of 90.36\% in comparison with 64.81\%  reported in \cite{Siamese}.
    %
    %
    \item We generalize the MAML-based diagnostic framework to identify real bearing damages using a model trained with artificial damages. Formulated as Artificial-to-Real tasks, the results indicate that the proposed framework has exceptional and robust generalization capabilities that are comparable to generalizing to new artificial faults.
\end{enumerate}

The rest of the paper is organized as follows. In Section II, we introduce some background knowledge and underlying principles of MAML. Next, in Section III, we present the architecture of the proposed MAML-based few-shot bearing fault diagnostic framework, with detailed descriptions on establishing the test environment and model implementation. Section IV presents results for case studies performed on few-shot bearing fault diagnosis using both the Case Western Reserve University (CWRU) bearing dataset \cite{CWRU} and the Paderborn dataset \cite{Paderborn, lessmeier2016condition}. Section V concludes the paper by highlighting the effectiveness of the proposed model in bearing fault diagnosis with limited data and discussing future work directions.
\section{Principle of Model-Agnostic Meta-Learning}
\subsection{Meta-Learning}
Meta learning is a general paradigm for few-shot learning. The objective of meta-learning is to learn a learning strategy to learn quickly on new tasks. In general, meta-learning algorithms involve two core processes: learning the transfer of knowledge across tasks and rapid adapting to new tasks \cite{ood_maml}. Therefore, meta learning is intrinsically well-suited to perform anomaly detection or fault diagnosis tasks on previously unseen tasks. Specifically, meta-learning is able to find good initial parameters of a neural network model, such that with only a few gradient steps, the initial parameters can be updated to produce a model that provides a good performance for a new fault category.

In standard supervised learning, the goal is to learn a function that maps from some input $x$, which might be an image, to the label of that image $y$, as illustrated in Eqn. (1). In supervised meta-learning, the idea is very similar except that now it's trying to map a training set $\mathcal{D}_{\text{train}}$ (with corresponding data and labels) and a test input $x$ to the label of that test input, as described in Eqn. (2). Essentially, the goal of meta-learning is to train a model that when exposed to a training set, performs well on a corresponding test set of that task.
\begin{align}
    & \textrm{Supervised learning: } f(x) \rightarrow y
    \\
    & \textrm{Supervised meta-learning: } f\left(\mathcal{D}_{\text{train}}, x\right) \rightarrow y 
\end{align}

The meta-training set $\mathcal{D}_{\text {train}}$ is typically designed to contain a collection of little datasets of different categories \cite{few_shot_optimization}. At meta-test time, the goal is to identify new categories of labels not previously seen in the training process using a limited amount of test data $x$. The way to accomplish this goal is to find a model that can accurately match this function $f\left(\mathcal{D}_{\text {train}}, x\right)$. Besides some successful deployment of sequence models in this effort such as recurrent neural networks \cite{few_shot_optimization} or temporal convolution networks \cite{TCN}, a very influential model was proposed by Finn et al. in \cite{MAML}, which was referred to as model-agnostic meta learning (MAML). 
\begin{figure*}[!t]
	\centering
	\includegraphics[height=5.0in]{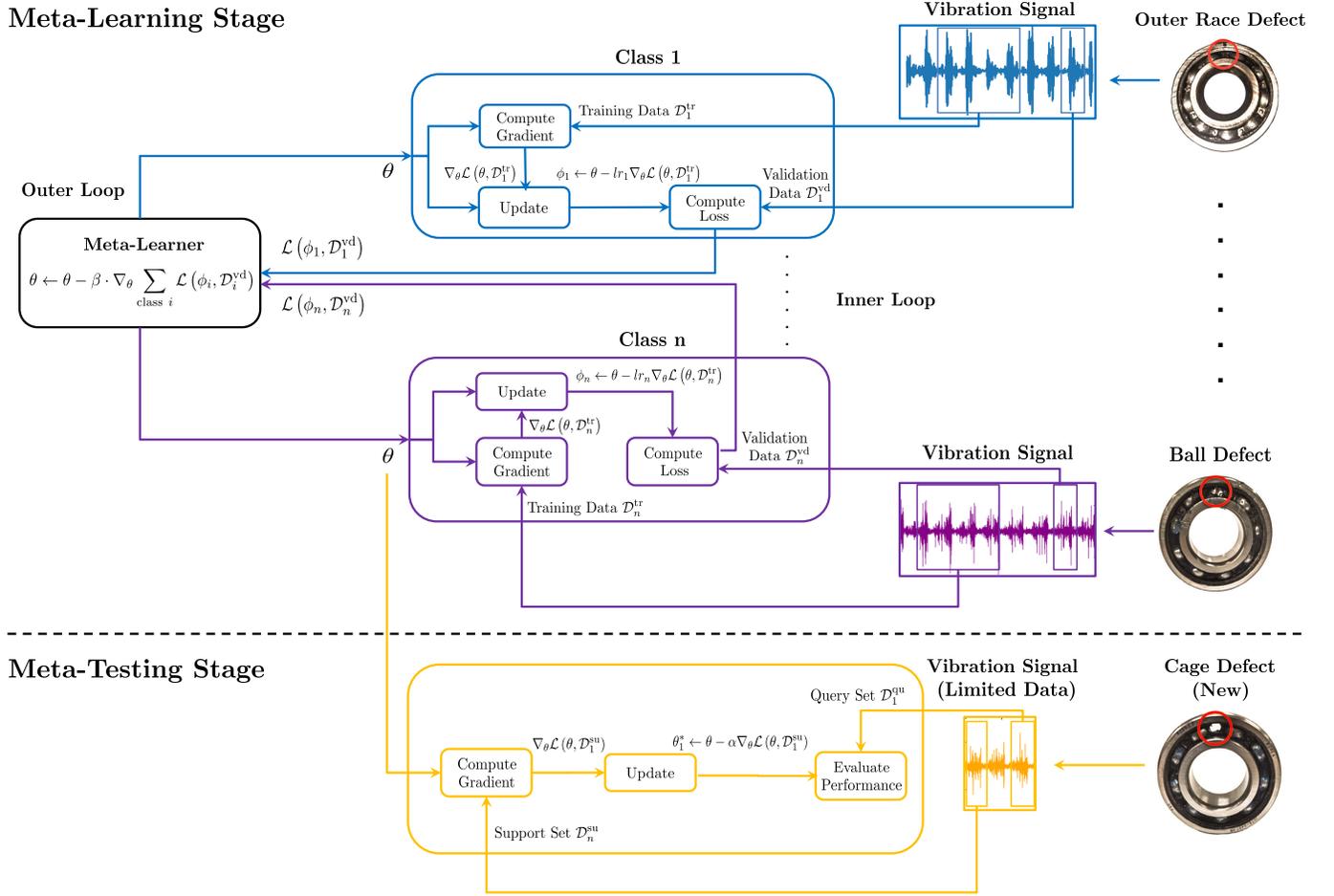}
	\caption{Flowchart of the meta-training stage of the proposed MAML-based few-shot bearing fault diagnostic framework.}
	\label{fig:flowchart}
\end{figure*}

\subsection{Meta-Agnostic Meta-Learning}
The MAML algorithm is model-agnostic. More specifically, it is agnostic both to the architecture of the neural network and also to the loss function. The backbone of MAML is to optimize for parameters that adapt quickly with gradient descent in two loops -- an inner loop and an outer loop. All of these features provide many flexibilities for MAML, making it applicable to both reinforcement learning problems that maximize the expected cumulative reward function, and to supervised learning problems that minimize a certain loss function (cross-entropy, mean-squared error, etc.). 

Performing a few-shot classification with MAML requires two stages -- a meta-training stage and a meta-testing stage, as shown in Fig. \ref{fig:flowchart}. Formally, we consider MAML as a neural network $f_{\theta}$ parameterized by $\theta$, which will be updated to $\phi_i$ using gradient descent when adapting to a new class $i$. During the meta-training stage, MAML operates in an inner loop and an outer loop. In the inner loop, MAML first computes the updated parameter vector $\phi_i$ for each class $i$ using training data $D^{tr}_i$, and then it evaluates the loss term on the validation data $D^{vd}_i$ sampled from the same class using the updated model parameters $\phi_i$. The evaluated loss for each class $i$ can be written as
\begin{equation}
\mathcal{L}\left(\phi_i, \mathcal{D}_{i}^{\mathrm{vd}}\right) = \mathcal{L}\left(\theta-\alpha \nabla_{\theta} \mathcal{L}\left(\theta, \mathcal{D}_{i}^{\mathrm{tr}}\right), \mathcal{D}_{i}^{\mathrm{vd}}\right)
\end{equation}
where $\phi_i \leftarrow \theta- \alpha \nabla_{\theta} \mathcal{L}\left(\theta, \mathcal{D}^{\mathrm{tr}}_i\right)$ is the updated model parameter for class $i$. The loss term is typically the cross-entropy loss for classification tasks. In the context of bearing fault diagnosis, as illustrated in Fig. \ref{fig:flowchart}, different classes can represent different types of bearing defects, such as inner/outer race defects, ball defects, cage defects, among others.

In the outer loop, MAML aggregates the per-task post-update losses $\mathcal{L}\left(\phi_i, \mathcal{D}_{i}^{\mathrm{vd}}\right)$ and performs a meta-gradient update on the original model parameter $\theta$ as
\begin{equation}
    \theta \leftarrow \theta- 
\beta\cdot\nabla_{\theta}\sum_{\mathrm{class}~i} \mathcal{L}\left(\phi_i, \mathcal{D}_{i}^{\mathrm{vd}}\right)
\end{equation}
where $\beta$ is the learning rate of the outer loop. At meta-test time, MAML is able to compute new model parameters based on only a few samples from unseen classes, and it uses the new model parameters to predict the label of a test sample from the same unseen class. 

In summary, the essential idea of MAML is trying to find parameters of a neural network that does not necessarily have the optimal performance for different classes of data provided at the meta-training stage, but  can quickly adapt to new (unseen) tasks.

%
%
\subsection{Meta-Agnostic Meta-Learning with Learnable Inner Loop Learning Rates}
As illustrated in Fig. \ref{fig:flowchart}, there is a learning rate $lr$ in MAML for its inner loop gradient update, which is assigned as a fixed number $\alpha$ in \cite{MAML} and is shared among different classes for all update steps. However, this fixed and shared inner loop learning rate $lr$ can often affect MAML's generalization capability and convergence speed \cite{Train_MAML}, and the process of tuning this hyper-parameter $lr$ for a specific dataset can often be costly and computationally intensive.

Therefore, a variant of MAML is proposed in \cite{Train_MAML} to automatically learn the inner loop learning rate $lr$. Specifically, it tries to learn different learning rates for each layer of the neural network and for each step through back-propagation. By doing this, the learning rate $lr$ becomes a vector that accounts for different learning rates for each layer of the neural network. With this learnable $\mathbf{lr}$ approach, elements in the learning rate vector $\mathbf{lr}$ can learn to decrease their values as the training progresses, which may help promote faster convergence and alleviate overfitting. Therefore, the revised form of Eqn. (3) to compute the loss term of each class can be written as
\begin{equation}
\mathcal{L}\left(\phi_i, \mathcal{D}_{i}^{\mathrm{vd}}\right) = \mathcal{L}\left(\theta- lr_i \nabla_{\theta} \mathcal{L}\left(\theta, \mathcal{D}_{i}^{\mathrm{tr}}\right), \mathcal{D}_{i}^{\mathrm{vd}}\right)
\end{equation}
\section{Few-Shot Bearing Fault Diagnostic Framework based on MAML}
\subsection{Proposed Few-Shot Bearing Fault Diagnostic Framework}
Few-shot classification is considered an instantiation of meta-learning in the field of supervised learning \cite{blog}. The standard few-shot learning is usually formulated as $N$-way $K$-shot problems, where $N$ is the number of new classes not seen in the meta-training process, while each class only has $K$ samples to train from. 
\begin{figure}
	\centering
	\includegraphics[height=1.9in]{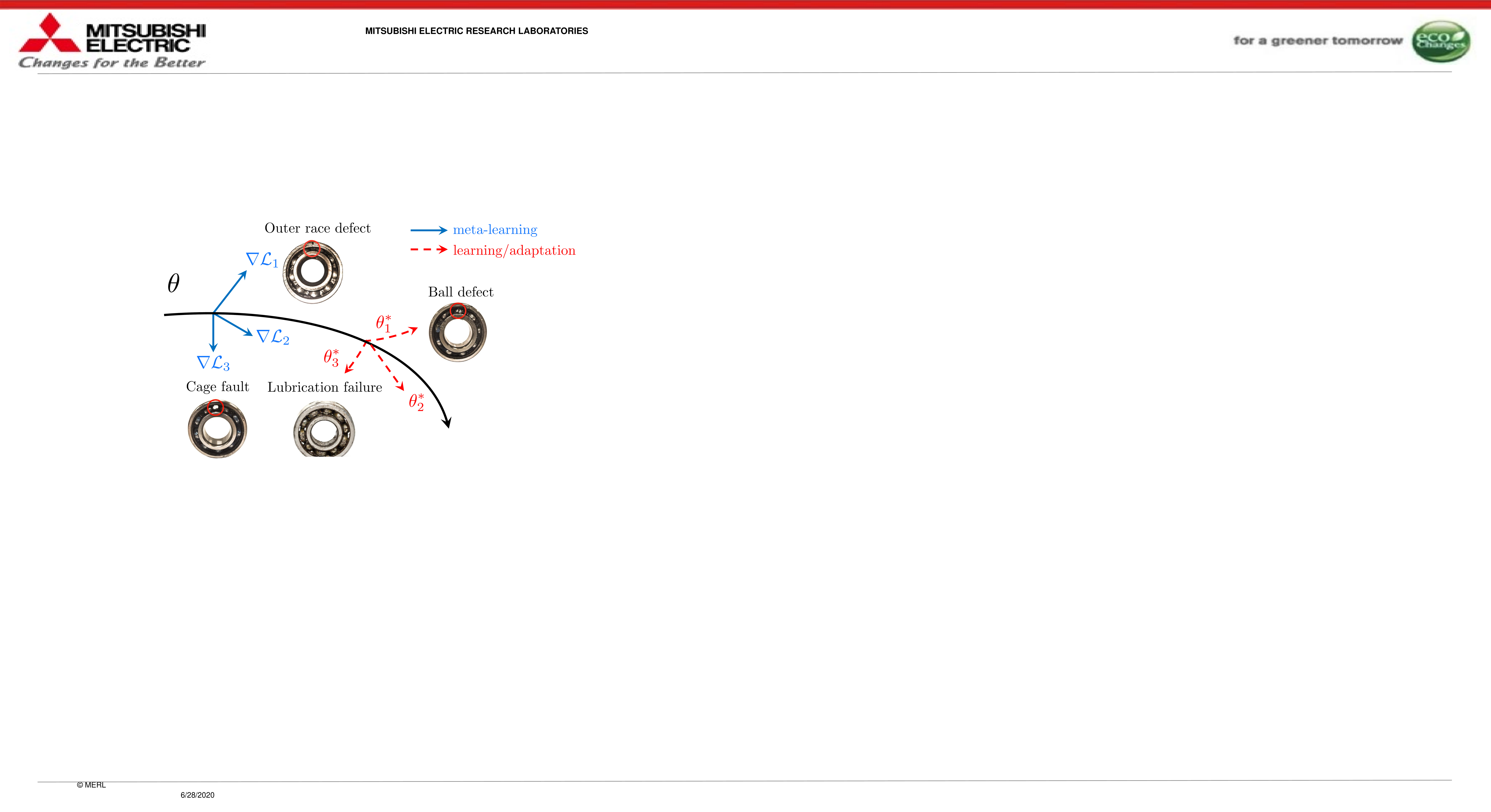}
	\caption{Illustration of MAML applied to the few-shot learning of bearing fault diagnosis}.
	\label{fig:MAML}
\end{figure}

The proposed few-shot bearing fault diagnostic framework based on MAML is illustrated in Fig. \ref{fig:MAML}. At the meta-training stage, we'll first optimize for a parameter set $\mathbf{\theta}$ of a neural network along the aggregated gradient descent direction of data from different bearing fault scenarios ($\nabla \mathcal{L}_{1}$, $ \nabla \mathcal{L}_{2}$, $\nabla \mathcal{L}_{3}$, etc.). As discussed in Section II, this parameter set $\mathbf{\theta}$ is optimized to achieve quick adaptation to new classes not previously seen at the meta-training stage, rather than achieving the optimal performance on classes it was directly trained on. 

For example, as shown in Fig. \ref{fig:MAML}, we can train the MAML-based diagnostic framework using data from bearing outer race defects and cage defects at different fault severity, and generalize it to detect new fault scenarios such as the ball defect and lubrication failure using a very small amount of data (e.g., $5$ samples). This problem will be formulated as a $2$-way $5$-shot few-shot learning setting that is well-suited for MAML.

With the proposed MAML-based few-shot bearing fault diagnostic framework, it is envisioned that we can mitigate both data scarcity and data imbalance issues discussed in \cite{Siamese} by adapting to these classes at the meta-testing stage, which can yield a satisfactory performance but only requires a limited amount of data. Additionally, another appealing application is to recognize naturally-evolved bearing defects using models that are only trained on data collected from artificial failures, since most bearing failures evolve slowly over time and it might take months if not years to collect a sufficient amount of data to train supervised learning-based fault classifiers.
\subsection{Objectives}
In the proposed MAML-based few-shot bearing fault diagnostic framework, the objective is to validate the performance of MAML on few-shot bearing fault diagnosis from the following aspects:
\begin{enumerate}
    \item[1)] \emph{Training Data Size}: Investigate the influence of training data size on the performance of MAML-based few-shot bearing fault diagnosis.
    \item[2)] \emph{New Artificially-Induced Bearing Failures}: Validate the performance of MAML to predict previously unseen artificial bearing faults in the laboratory environment.
    \item[3)] \emph{New Realistic Bearing Failures with Accelerated Aging}: Explore the generalization capability of MAML to predict real bearing failures with accelerated lifetime tests using data collected from artificially-damaged bearings.
\end{enumerate}

Both 1) and 2) have been investigated in \cite{Siamese} using the Siamese Network-based few-shot learning method on the CWRU dataset. In order to perform a fair comparison, we strive to keep the test environment consistent with the benchmark study by also leveraging the CWRU dataset and assigning the same fault labels. More details regarding the CWRU dataset can be found in their website \cite{CWRU}. 
\begin{table}[!t]
\begin{center}
   \caption{Different categories of bearing failures selected from the CWRU dataset \cite{CWRU}}
{\footnotesize
\begin{tabular}{ccc}
\toprule
$\mathrm{Class~Label}$ & $\mathrm{Fault~Location}$ & $\mathrm{Fault~Diameter~(mils)}$        \\
\midrule
1                    & {$\mathrm{Healthy}$ }    & $0$      \\
\midrule
2                    & $\mathrm{Ball}$         & $0.007$     \\
3                    & $\mathrm{Ball}$         & $0.014$     \\
4                    & $\mathrm{Ball}$         & $0.021$     \\
\midrule
5                    & $\mathrm{Inner~Race}$   & $0.007$     \\   
6                    & $\mathrm{Inner~Race}$   & $0.014$     \\    
7                    & $\mathrm{Inner~Race}$   & $0.021$     \\ 
\midrule
8                    & $\mathrm{Outer~Race}$   & $0.007$     \\ 
9                    & $\mathrm{Outer~Race}$   & $0.014$     \\
10                   & $\mathrm{Outer~Race}$   & $0.021$     \\
\bottomrule
\end{tabular}} 
\end{center}

\begin{footnotesize}
\hspace{0.3in} Each class contains fault data collected at 3 different operating \\
\hspace*{0.3in} conditions: 1) 1 hp load at 1772 rpm; 2) 2 hp load at 1750 rpm; and 
\hspace*{0.3in} 3) 3 hp load at 1730 rpm.
\end{footnotesize}
\label{tab:CWRU}
\end{table}

A list of all 10 fault scenarios is presented in TABLE \ref{tab:CWRU}. Specifically, different classes are identified based on the location and size of a bearing defect, rather than its operating speed and loading condition. We also adopt the same data segmentation method as \cite{Siamese}, in which each data segment consists of $2048\times 2$ data points that are sampled at 12 kHz from both accelerometers at the Fan end and the Load end. After performing the aforementioned classification and data segmentation strategies, the entire CWRU dataset is partitioned into 10 classes, with each class having 1,980 data segments. 

\begin{figure}[!t]
\centering
\includegraphics[width=3.4in]{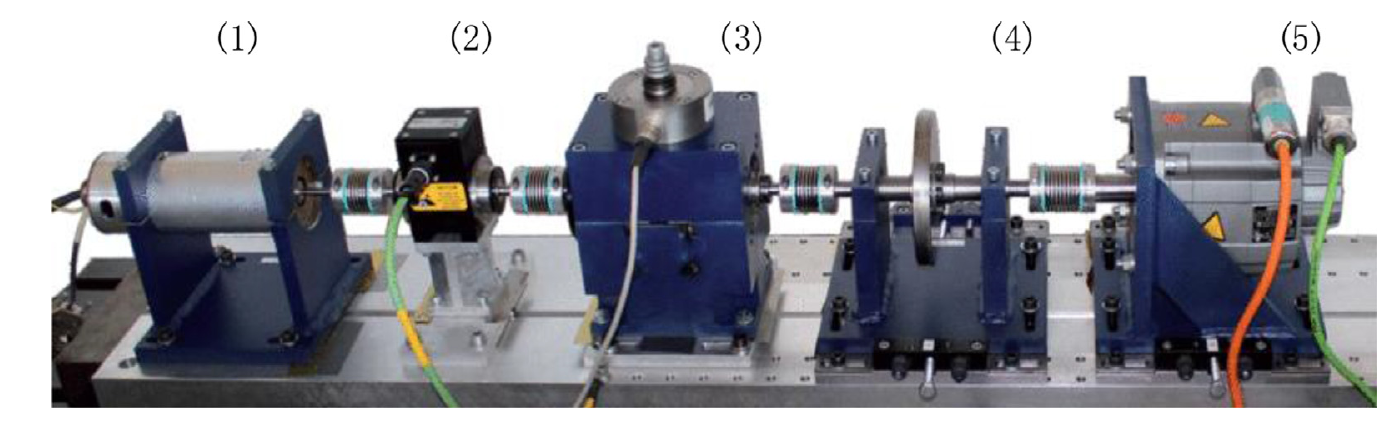}
\caption{Modular test rig collecting the Paderborn bearing dataset consisting of (1) an electric motor, 
(2) a torque-measurement shaft, (3) a rolling bearing test module, (4) a flywheel, and (5) a load motor \cite{Paderborn}.}
\label{fig:Paderborn}
\end{figure}

To further investigate the generalization capability of MAML in predicting real bearing failures as described in 3), we'll also apply the proposed MAML-based few-shot learning framework to the Paderborn dataset \cite{Paderborn}, since the CWRU dataset only contains artificially-induced defects. The Paderborn dataset includes data of 32 bearings under test, and among them, 6 are normal ones, 12 are with artificially-induced damages, and 14 are with real damages caused by accelerated aging tests. There are only inner and outer raceway defects present for both artificial and real bearing failures, while damages at the rolling elements were not observed. Each of the 32 fault categories contains data collected at 4 different operating conditions with different combinations of rotational speed, load torque, and radial force. The experimental setup used to collect the Paderborn dataset is shown in Fig. \ref{fig:Paderborn}.
\begin{table}[!t]
\caption{Different categories of bearing failures selected from the Paderborn dataset}
\begin{center}
\resizebox{\linewidth}{!}{
\begin{tabular}{ccccc}
\toprule
$\mathrm{Label}$ & $\mathrm{Fault~Location}$ & $\mathrm{Cause~of~Failure}$    & $\mathrm{Severity}$ & $\mathrm{Code}$   \\
\midrule
1                    & {$\mathrm{Healthy}$ }   & $\mathrm{N/A}$   & 0    & K001  \\
\midrule
2                    & $\mathrm{Outer~Race}$   & $\mathrm{EDM}^*$ & 1  & KA01    \\
3                    & $\mathrm{Outer~Race}$   & $\mathrm{EE^\ddagger}$ & 2 & KA03   \\
4                    & $\mathrm{Outer~Race}$   & $\mathrm{Drilling}$ & 1 & KA07   \\
5                    & $\mathrm{Inner~Race}$   & $\mathrm{EDM}$ & 2 & KI01   \\  
6                    & $\mathrm{Inner~Race}$   & $\mathrm{EE^\ddagger}$  & 1 & KI03   \\  
7                    & $\mathrm{Inner~Race}$   & $\mathrm{EE^\ddagger}$  & 2 & KI07   \\ 
\midrule
8                    & $\mathrm{Outer~Race}$   & $\mathrm{Pitting}$ & 1 & KA04   \\ 
9                    & $\mathrm{Outer~Race}$   & $\mathrm{PD}^\dagger$ & 1 & KA15   \\
10                   & $\mathrm{Outer~Race}$   & $\mathrm{Pitting}$ & 2  & KA16   \\
11                   & $\mathrm{Inner~Race}$   & $\mathrm{Pitting}$ & 1 & KI04   \\ 
12                   & $\mathrm{Inner~Race}$   & $\mathrm{Pitting}$ & 3 & KI16   \\
13                   & $\mathrm{Inner~Race}$   & $\mathrm{Pitting}$ & 2 & KI18   \\
\bottomrule
\end{tabular}}
\end{center}
\begin{footnotesize}
\hspace{0.1in}$^*$EDM: Electrical discharge machining.\\
\hspace*{0.1in}$^\ddagger$EE: Electric engraver.\\
\hspace*{0.1in}$^\dagger$PD: Plastic deform.
\end{footnotesize}
\label{tab:Paderborn}
\end{table}

For the Paderborn bearing dataset, we also select 13 representative classes from the total 32 classes, with 1 at the healthy condition, 6 have manually-initiated bearing defects, and the rest have real bearing failures resulting from accelerated lifetime tests. The selected classes can be distinguished based on their fault locations, causes of failure, and fault severity. The threshold values used to determine different levels of fault severity have been discussed in detail in \cite{Paderborn_paper}, where level 1 corresponds to a defect length smaller than 2 mm, level 2 corresponds to defect lengths between 2 to 4.5 mm, and level 3 corresponds to 4.5 to 13.5 mm.

A complete list of the selected classes is presented in TABLE \ref{tab:Paderborn}, and the goal is to successfully identify real bearing failures (categories 8 to 13) using the healthy and artificial fault data (categories 1 to 7). The rest of the data segmentation process is consistent with that performed on the CWRU data.
\subsection{Model Implementations}
The model follows the same architecture as the embedding function used by \cite{vinyals2016matching}, which has 4 modules with a $3 \times 3$ convolutions and 64 filters, followed by batch normalization, a ReLU nonlinearity, and $2 \times 2$ max-pooling. The bearing vibration signals are sampled with a dimension of 4096 and converted to $64 \times 64$, and the last layer is fed into a softmax layer. For $N$-way, $K$-shot classification tasks, each gradient is computed using a batch size of $NK$ examples.

The $N$-way convolutional model is trained with 1 gradient step and a meta batch-size of 25 tasks. For MAML with a fixed learning rate, the learning coefficient is chosen as $\alpha = 0.01$. For MAML with a learnable inner loop learning rate $lr$, the initial value is also kept at 0.01, and it will be optimized with the training step \cite{Train_MAML}. We used a meta batch-size of 1 task for both 1-shot and 5-shot testing. All models were trained for 1500 iterations.

\section{Experimental Results}
In this section, we seek to validate the performance of the MAML-based few-shot bearing fault diagnostic framework. As discussed in Section III, We'll specifically investigate its performance with different training data size and different unseen fault categories using the CWRU dataset. Additionally, we'll also leverage the Paderborn dataset to predict naturally-evolved bearing failures using data collected from artificially-damaged bearings. The results of the proposed MAML-based few-shot classifier will be compared against those obtained using the Siamese Network in \cite{Siamese}, and we strive to keep their test scenarios as consistent as possible by leveraging the open-source code provided in \cite{Siamese}.

\subsection{Ablation Study}

\subsubsection{Influence of the Inner Loop Learning Rate $l_r$ and the Optimizer Type}
To start with, we evaluate the influence of the inner loop learning rate $l_r$ and the optimizer type (e.g., Adam, RMSprop, and stochastic gradient descent (SGD)) on the performance of the proposed MAML-based few-shot bearing fault diagnostic framework. We conducted a series of comparison experiments by setting the last 3 classes of TABLE \ref{tab:CWRU} as the test dataset. In this way, data related to the outer bearing defect is only contained in the meta-testing data rather than the meta-training data. Then this setting is formulated as a $3$-way few-shot learning problem.  We select 9 data samples per class at the meta-training stage, and in the end, we also perform a 5-shot case study at the meta-testing stage.

\begin{table}[!t]
\caption{$3$-Way $5$-Shot classification results predicting the unseen outer race defect using different values of inner loop learning rate $l_r$ and different types of optimizers.}
\resizebox{\linewidth}{!}{
\begin{tabular}{lccc}
\toprule
Optimizers                  & $l_r = 0.001$ & $l_r = 0.01$ & $l_r=0.1$ \\
\midrule
Adam                  &       $85.70 \pm 1.66\%$           &   $\mathbf{90.36 \pm 1.38}\%$        &    $86.80 \pm 1.31\%$          \\
RMSprop           &    $87.36 \pm 1.63\%$        &    $85.70 \pm 1.66\%$       &   $82.16 \pm 2.58\%$     \\
SGD &    $82.68 \pm 1.46\%$        &    $79.45 \pm 1.22\%$   &  $76.26 \pm 3.18\%$
\\
\bottomrule
\end{tabular}}
\label{tab:ablation}
\end{table}

The meta-testing stage is repeated 10 times, and the mean accuracy and standard deviation are presented in TABLE \ref{tab:ablation}. It can be observed that in the proposed $3$-way $5$-shot learning setting with the unseen outer race defect during meta-training, the proposed MAML framework with learnable $\mathbf{lr}$ can achieve the optimal performance with the Adam optimizer and an inner loop learning rate of $l_r=0.01$. Specifically, the margin of mean accuracy is at least 3\% when compared with other cases.

\subsubsection{Influence of Training Data Size}
\begin{table*}[!t]
\centering
\caption{$3$-Way $K$-Shot classification results predicting the unseen outer race defect using different numbers of training samples.}
\label{table:training}
\begin{tabular}{lccccc}
\toprule
{Number of Training Samples Per Class}  & $4$ & $6$ &  $9$ &  $12$  \\
\midrule
Siamese Network $1$-shot \cite{Siamese} & $53.22 \pm 4.38\%$ & $59.69 \pm 6.25\%$ & $63.79 \pm 4.45\%$ & $66.08 \pm 5.56\%$ \\
Siamese Network $5$-shot \cite{Siamese} & $53.44 \pm 4.93\%$ & $59.88 \pm 6.04\%$ & $64.81 \pm 6.32\%$ & $66.57 \pm 5.74\%$ \\
\textbf{MAML (learnable $\mathbf{lr}$)} $1$-shot & $78.45 \pm 2.94\%$ & $84.71 \pm 2.53\%$ & $88.39 \pm 2.95\%$ & $96.29 \pm 1.66\%$ \\
\textbf{MAML (learnable $\mathbf{lr}$)} $5$-shot & $\mathbf{78.54 \pm 2.41}\%$ & $\mathbf{86.79 \pm 1.21}\%$ & $\mathbf{90.36 \pm 1.38}\%$ & $\mathbf{99.77 \pm 0.20}\%$ \\
\bottomrule
\end{tabular}
\end{table*}

We then evaluate the influence of training data size with test setup similar to the previous section as a $3$-way few-shot learning problem. The optimal combination of the Adam optimizer and $l_r=0.01$ is also selected based on results presented in TABLE \ref{tab:ablation}.

The training data size per class is selected to be 4, 6, 9, and 12, and these samples are randomly chosen from the 1980 segments in each class in TABLE \ref{tab:CWRU}. The results obtained from a Siamese Network-based few-shot bearing diagnostic framework proposed in \cite{Siamese} are also provided as a benchmark. The original results in \cite{Siamese} only include the case of 9 training samples per class, and we obtained the results of the remaining cases using the open-source code provided in \cite{Siamese}. For each size of the training set, we repeated the algorithm training and testing stages 5 times to overcome the randomness of both algorithms. The average accuracy and the standard deviation are presented in TABLE \ref{table:training} with different numbers of training samples per class.

It can be observed that in the proposed $3$-way few-shot learning setting, the proposed MAML-based framework with learnable $\mathbf{lr}$ consistently outperforms the benchmark Siamese network by 20\% to 30\%. In addition, the proposed method can achieve an average accuracy of 99.77\% with $5$-shot learning and 12 training samples per class. While this result obtained using 12 samples per class is indeed satisfactory, we decide to use 9 samples per class for later experiments. This is because we seek to further investigate the performance of the proposed framework predicting different unseen fault types, and a near-perfect accuracy may overshadow other intrinsic processes undergoing in MAML. Additionally, we can make the test environment more consistent with the benchmark study \cite{Siamese}, which also presents its results using 9 training samples per class.
\subsection{Predicting New Artificially-Induced Bearing Defects}

The experiments performed in this section seeks to investigate MAML's performance to predict artificially-induced bearing faults that are not seen at the meta-training stage. We also extract data from the CWRU dataset to test $1$-way to $5$-way classifications with 1 and 5 shots. A total of $10$ rounds of experiments are performed to compare with the benchmark in \cite{Siamese}. Since \cite{Siamese} only provides the $1$-way to $3$-way results, the 4-way and 5-way results are also obtained using the open-source code in \cite{Siamese}. We also followed the order of labels presented in TABLE \ref{tab:CWRU}, thus a $5$-way classification indicates we are deploying data with class labels 1 to 5 as the meta-training data, and the rest will serve as the meta-testing data.
\begin{table*}[!t]
\centering
\label{table:few_shot}
\caption{$N$-Way $K$-Shot classification results predicting new types of artificial bearing faults (9 training samples per class).}
\resizebox{5.0in}{!}{
\begin{tabular}{lcccc}
\toprule
\multirow{2}{*}{$N$-way Accuracy} & \multicolumn{2}{c}{$5$-way Accuracy} & \multicolumn{2}{c}{$4$-way Accuracy} \\ 
\cmidrule(lr){2-5}
\cmidrule(lr){4-5}
 & $1$-shot &  $5$-shot &  $1$-shot &  $5$-shot \\
\midrule
Siamese-nets \cite{Siamese} & $46.95 \pm 4.23\%$ & $52.80 \pm 5.67\%$ & $56.12 \pm 2.29\%$ & $60.92 \pm 2.47\%$ \\
MAML (fixed $lr$) & $56.31 \pm 2.07\%$ & $59.86 \pm 1.04\%$ & $66.91 \pm 3.63\%$ & $70.68 \pm 2.39\%$ \\
\textbf{MAML (learnable $\mathbf{lr}$)} & $\mathbf{76.55 \pm 1.57}\%$ & $\mathbf{79.98 \pm 1.17}\%$ & $\mathbf{78.31 \pm 2.99}\%$ & $\mathbf{83.45 \pm 2.26}\%$ \\
\bottomrule
 \multicolumn{5}{c}{} 
\end{tabular}}
\begin{tabular}{lcccccc}
\toprule
\multirow{2}{*}{$N$-way Accuracy} & \multicolumn{2}{c}{$3$-way Accuracy} & \multicolumn{2}{c}{$2$-way Accuracy} & \multicolumn{2}{c}{$1$-way Accuracy} \\ 
\cmidrule(lr){2-3}
\cmidrule(lr){4-5}
\cmidrule(lr){6-7}
{ }  &  $1$-shot &  $5$-shot &  $1$-shot &  $5$-shot &  $1$-shot &  $5$-shot\\
\midrule
Siamese-nets \cite{Siamese} & $63.79 \pm 4.45\%$ & $64.81 \pm 6.32\%$ & $81.23 \pm 3.36\%$ & $83.04 \pm 2.42\%$ & $87.04 \pm 3.73\%$ & $88.2 \pm 3.56\%$ \\
MAML (fixed $lr$) & $82.73 \pm 3.12\%$ & $85.53 \pm 1.07\%$ & $100\%$ & $100\%$ & $100\%$ & $100\%$ \\
\textbf{MAML (learnable $\mathbf{lr}$)} & $\mathbf{88.39 \pm 2.95}\%$ & $\mathbf{90.36 \pm 1.38}\%$ & $\mathbf{100}\%$ & $\mathbf{100}\%$ & $\mathbf{100}\%$ & $\mathbf{100}\%$ \\
\bottomrule
 \multicolumn{7}{c}{} 
\end{tabular}
\end{table*}

\begin{table*}[!t]
\centering
\caption{$3$-way $K$-shot classification results on the CWRU dataset predicting new types of artificial bearing faults.}
\label{table:new_fault}
\resizebox{5.0in}{!}{
\begin{tabular}{lcccc}
\toprule
\multirow{2}{*}{New Fault Category} & \multicolumn{2}{c}{Outer Race Defect} & \multicolumn{2}{c}{Inner Race Defect}  \\ 
\cmidrule(lr){2-3}
\cmidrule(lr){4-5}
{ }  &  $1$-shot &  $5$-shot &  $1$-shot &  $5$-shot \\
\midrule
Siamese-nets \cite{Siamese} & $63.79 \pm 4.45\%$ & $64.81 \pm 6.32\%$ & $60.82 \pm 4.03\%$ & $62.05 \pm 2.91\%$ \\
MAML (fixed $lr$) & $82.73 \pm 3.12\%$ & $85.53 \pm 1.07\%$ & $80.72 \pm 2.60\%$ & $81.21 \pm 1.51\%$ \\
\textbf{MAML (learnable $\mathbf{lr}$)} & $\mathbf{88.39 \pm 2.95}\%$ & $\mathbf{90.36 \pm 1.38}\%$ & $\mathbf{88.62 \pm 4.03}\%$ & $\mathbf{91.55 \pm 1.39}\%$ \\
\bottomrule
 \multicolumn{5}{c}{} 
\end{tabular}}
\begin{tabular}{lcccc}
\toprule
\multirow{2}{*}{New Fault Category} & \multicolumn{2}{c}{Ball Defect} & \multicolumn{2}{c}{No (New Classes $(4, 7, 10)$ only with Different Defect Sizes)}  \\ 
\cmidrule(lr){2-3}
\cmidrule(lr){4-5}
{ }  &  $1$-shot &  $5$-shot &  $1$-shot &  $5$-shot \\
\midrule
Siamese-nets \cite{Siamese} & $63.79 \pm 4.45\%$ & $64.81 \pm 6.32\%$ & $69.82 \pm 2.34\%$ & $72.06 \pm 3.10\%$ \\
MAML (fixed $lr$) & $78.53 \pm 5.12\%$ & $81.57 \pm 3.72\%$ & $89.41 \pm 4.49\%$ & $92.08 \pm 1.62\%$ \\
\textbf{MAML (learnable $\mathbf{lr}$)} & $\mathbf{88.62 \pm 4.03}\%$ & $\mathbf{88.86 \pm 1.83}\%$ & $\mathbf{95.23 \pm 1.19\%}\%$ & $\mathbf{99.49 \pm 0.41\%}\%$ \\
\bottomrule
\end{tabular}
\end{table*}
\begin{figure}
	\centering
	\includegraphics[height=2.2in]{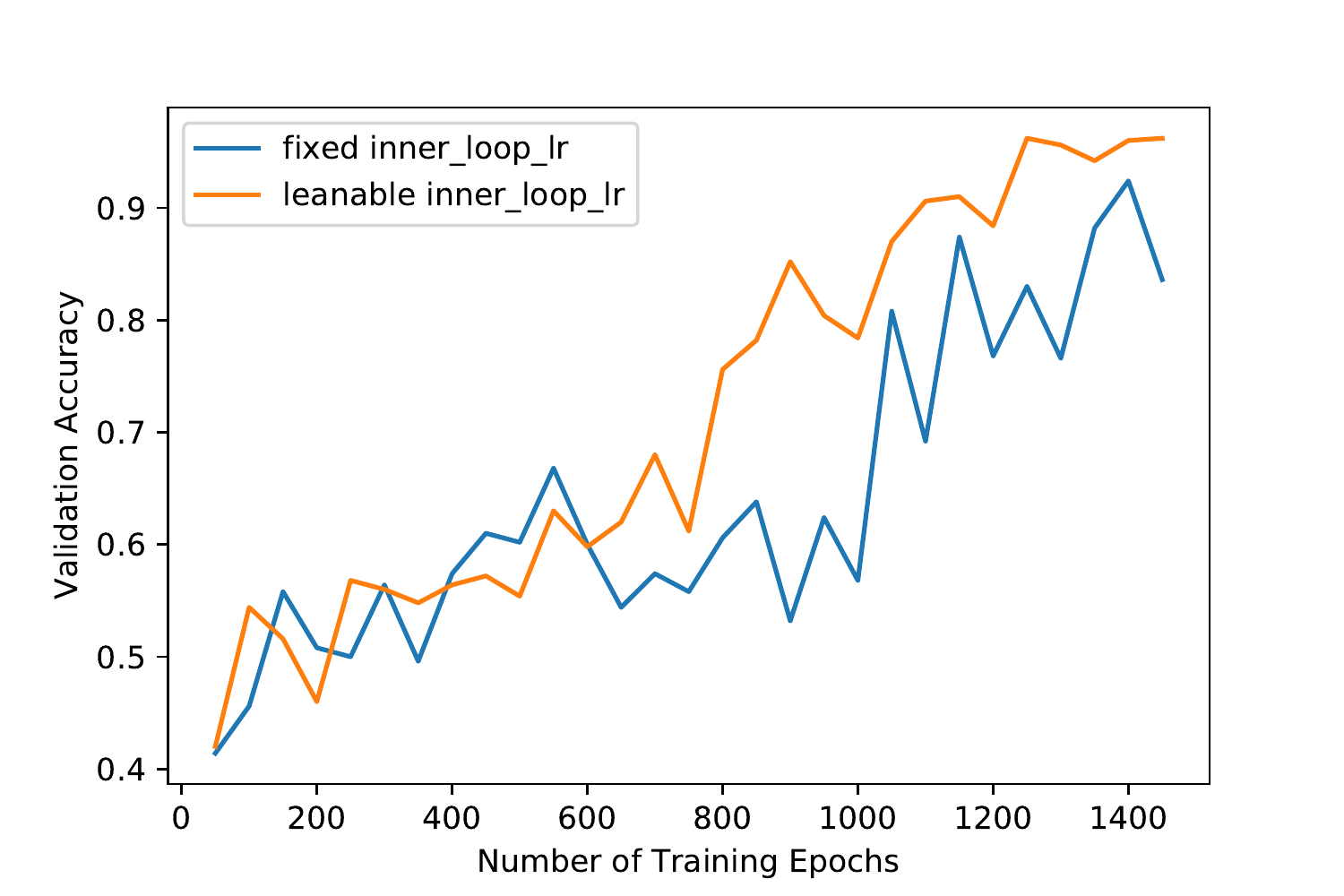}
	\caption{Diagram of model-agnostic meta-learning algorithm applied to few-shot learning of bearing fault diagnosis.}
	\label{fig:Learnable_rate}
\end{figure}

By randomly selecting 9 samples from the 1,980 available ones for each class, the complete results on $N$-way $K$-shot classification are presented in TABLE IV. It can be observed that MAML with a fixed $lr$ is able to achieve 10\% to 20\% enhancement in average accuracy when compared with the benchmark Siamese Network, while MAML with learnable $\mathbf{lr}$ is able to achieve even larger improvements ranging from 20\% to 30\%. A comparison study for MAML with fixed and learnable $\mathbf{lr}$ is illustrated in Fig. \ref{fig:Learnable_rate}, where the validation accuracy of the learnable inner update $lr$  consistently outperforms, and is also more stable than the fixed $lr$ case after 600 training epochs. This observation can be interpreted in such a way that the learnable $\mathbf{lr}$ can learn to decrease the learning rates with larger training epochs and getting closer to the local optimum, which may help alleviate overfitting and promote faster convergence \cite{Train_MAML}.
\begin{table*}[!t]
\centering
\caption{$N$-way $K$-shot classification results predicting new types of realistic bearing faults (9 training samples per class).}
\label{table:real}
\begin{tabular}{lcccccc}
\toprule
\multirow{2}{*}{$N$-way Accuracy} & \multicolumn{2}{c}{$6$-way Accuracy} & \multicolumn{2}{c}{$5$-way Accuracy} & \multicolumn{2}{c}{$4$-way Accuracy} \\ 
\cmidrule(lr){2-3}
\cmidrule(lr){4-5}
\cmidrule(lr){6-7}
{ }  &  $1$-shot &  $5$-shot &  $1$-shot &  $5$-shot &  $1$-shot &  $5$-shot\\
\midrule
%
MAML (fixed $lr$) & $44.36 \pm 4.02\%$ & $47.30 \pm 2.29\%$ & $55.51 \pm 3.02\%$ & $53.10 \pm 8.4\%$ & $74.93 \pm 5.05\%$ & $76.79 \pm 3.09\%$ \\
\textbf{MAML (learnable $\mathbf{lr}$)} & $\mathbf{55.21 \pm 3.01}\%$ & $\mathbf{62.58 \pm 2.78}\%$ & $\mathbf{73.63 \pm 3.75}\%$ & $\mathbf{78.15 \pm 2.76}\%$ & $\mathbf{75.04 \pm 3.51}\%$ & $\mathbf{84.62 \pm 1.10}\%$ \\
\bottomrule
 \multicolumn{7}{c}{} 
\end{tabular}
\resizebox{\linewidth}{!}{
\begin{tabular}{lcccccc}
\toprule
\multirow{2}{*}{$N$-way Accuracy} & \multicolumn{2}{c}{$3$-way Accuracy} & \multicolumn{2}{c}{$2$-way Accuracy} & \multicolumn{2}{c}{$1$-way Accuracy} \\ 
\cmidrule(lr){2-3}
\cmidrule(lr){4-5}
\cmidrule(lr){6-7}
{ }  &  $1$-shot &  $5$-shot &  $1$-shot &  $5$-shot &  $1$-shot &  $5$-shot\\
\midrule
%
MAML (fixed $lr$) & $82.73 \pm 3.12\%$ & $85.53 \pm 1.07\%$ & $97.68 \pm 2.18\%$ & $98.19 \pm 1.98\%$ & $98.77 \pm 2.06\%$ & $100\%$ \\
\textbf{MAML (learnable $\mathbf{lr}$)} & $\mathbf{85.57 \pm 2.85}\%$ & $\mathbf{97.90 \pm 1.50}\%$ & $\mathbf{97.85 \pm 0.61\%}$ & $\mathbf{100}\%$ & $\mathbf{100}\%$ & $\mathbf{100}\%$ \\
\bottomrule
 \multicolumn{7}{c}{} 
\end{tabular}
}
\end{table*}
\begin{figure*}
	\centering
	\includegraphics[height=2.0in]{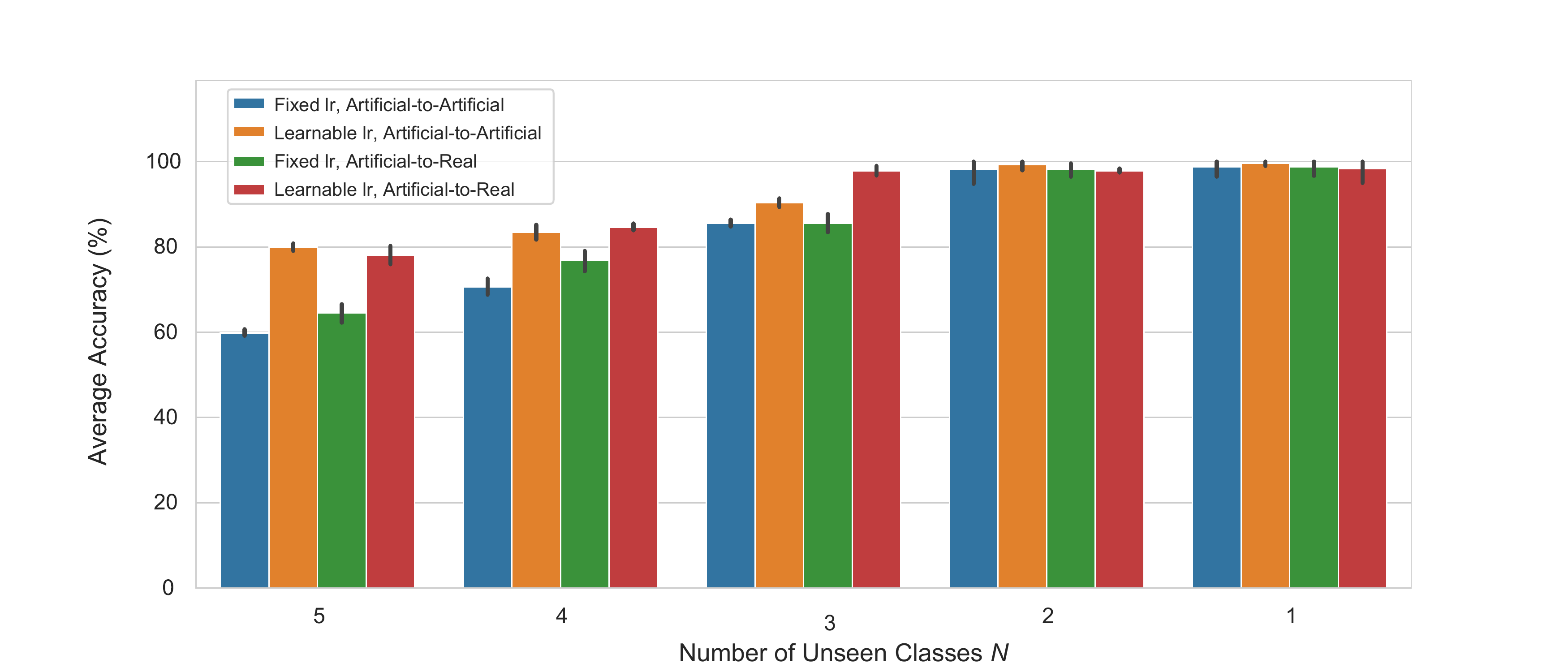}
	\caption{Comparison of the generalization capability of $N$-way $5$-Shot MAML between Artificial-to-Artificial and Artificial-to-Real.}
	\label{fig:artificial-to-real}
\end{figure*}

Besides examining the performance of MAML in $N$-way classification, it is also of practical value to investigate its performance at predicting different types of bearing faults not previously seen at the meta-training stage. An example of an unseen outer race defect is in fact the $3$-way example demonstrated in TABLE IV, where the term ``$3$-way'' corresponds to classes 8, 9, 10 in TABLE \ref{tab:CWRU}, and all of them contain data with bearing outer race defects that are not included in the meta-training data.

The results of MAML predicting other types of artificial bearing failures previously unseen at the meta-training stage are presented in TABLE \ref{table:new_fault}, where the meta-testing data for the inner race fault are sampled from classes 5, 6, 7, and data for the ball failure are sampled from classes 2, 3, 4, as listed in TABLE \ref{tab:CWRU}. It is shown in these 3 cases that MAML consistently outperforms the benchmark Siamese Network by around 20\% with a fixed $lr$ and by 25\% with a learnable $\mathbf{lr}$, when using the same training and test set over 1,500 training epochs. The $5$-shot accuracy with learnable $\mathbf{lr}$ is around 90\% for all 3 cases, validating the effectiveness of MAML predicting new types of bearing defects using a limited amount of data.

Another interesting case study is to predict bearing faults with higher levels of fault severity, or class (4, 7, 10) shown in TABLE \ref{tab:CWRU}, while their fault types are not new to MAML, since the training data (classes 1, 2, 3, 5, 6, 8, 9) already covers all of the fault categories. This setting can still be formulated as a $3$-way few-shot learning problem, and we are able to obtain around 10\% improvement in average accuracy when compared with the earlier 3 cases with completely new fault categories. This finding can be explained as the task distribution of the same type of fault with different levels of fault severity is closer than the distribution of different fault types. This observation leads to a potential guideline of selecting the meta-training data from as many fault categories as possible when employing the MAML-based few-shot bearing fault diagnostic framework.
\subsection{Predicting New Realistic Bearing Defects}
\begin{table}[!t]
\caption{Different categories of bearing failures selected from the Paderborn dataset to perform artificial-to-real few-shot learning \cite{Few_shot_transfer}}
\begin{center}
\resizebox{\linewidth}{!}{
\begin{tabular}{lcccc}
\toprule
\multicolumn{5}{c}{Artificial damages (source domain)} \\
\midrule
$\mathrm{Bearing}$ & $\mathrm{Fault}$ & $\mathrm{Cause~of}$    & \multirow{2}{*}{$\mathrm{Severity}$} &  $\mathrm{Characteristics}$   \\
$\mathrm{Code}$ & $\mathrm{Location}$ &  $\mathrm{Failure}$   &  & $\mathrm{of~Damage}$   \\
\midrule
KA01               & $\mathrm{Outer~Race}$   & $\mathrm{EDM}^*$ & 1  &  Punctual   \\
KA03                    & $\mathrm{Outer~Race}$   & $\mathrm{EE^\ddagger}$ & 2 &  Punctual  \\
KA05                    & $\mathrm{Outer~Race}$   & $\mathrm{EE^\ddagger}$ & 1 &  Punctual  \\
KA07                    & $\mathrm{Outer~Race}$   & $\mathrm{Drilling}$ & 1 & Punctual   \\  
KA08                    & $\mathrm{Inner~Race}$   & $\mathrm{Drilling}$  & 2 & Punctual   \\  
KI01                    & $\mathrm{Inner~Race}$   & $\mathrm{EDM}^*$  & 1 & Punctual   \\ 
KI03                    & $\mathrm{Inner~Race}$   & $EE^\ddagger$  & 1 & Punctual   \\ 
KI05                    & $\mathrm{Inner~Race}$   & $EE^\ddagger$  & 2 & Punctual   \\ 
\midrule
\multicolumn{5}{c}{Healthy \& Real damages (target domain)} \\
\midrule
K001                    & {$\mathrm{Healthy}$ }   & $\mathrm{N/A}$   & 0    &  $\mathrm{N/A}$ \\
KA04                    & $\mathrm{Outer~Race}$   & $\mathrm{Pitting}$ & 1 &  Punctual  \\ 
KB23                    & $\mathrm{Inner+(Outer)}$   & $\mathrm{Pitting}$ & 2 & Punctual   \\ 
KB27                    & $\mathrm{Inner+(Outer)}$   & $\mathrm{^\dagger PD}$ & 1 & Distributed   \\
KI04                    & $\mathrm{Inner~Race}$   & $\mathrm{Pitting}$  & 1 &  Punctual  \\ 
\bottomrule
\end{tabular}}
\end{center}
\begin{footnotesize}
\hspace{0.1in}$^*$EDM: Electrical discharge machining.\\
\hspace*{0.1in}$^\ddagger$EE: Electric engraver.\\
\hspace*{0.1in}$^\dagger$PD: Plastic deform.
\end{footnotesize}
\label{tab:Paderborn_real}
\end{table}

\begin{table*}[!t]
\centering
    \caption{Comparison of MAML with Different Benchmark Algorithms in the Artificial-to-Real Transfer Task.}
\begin{tabular}{lccccc}
\toprule
Algorithms  & $1$-shot  & $3$-shot   & $5$-shot   & $10$-shot  & Overall  \\
\midrule
Feature Transfer Net \cite{feature_transfer} & $87.69\pm 3.25\%$ & $94.12\pm 2.50\%$ & $96.29\pm 0.74\%$ & $\mathbf{97.48\pm 0.32\%}$& $93.90\%$ \\
Unfrozen 1 Fine-tuning Net \cite{Few_shot_transfer}  & ${92.16\pm 4.88\%}$  & $92.83\pm 3.61\%$ & $95.67\pm 2.44\%$ & $96.01\pm 1.27\%$ & $94.02\%$ \\
Unfrozen 2 Fine-tuning Net \cite{Few_shot_transfer} & $90.96\pm 4.88\%$ & $94.04\pm 4.55\%$ & $95.48\pm 2.78\%$ & $96.01\pm 1.57\%$ & $94.12\%$  \\
Unfrozen 3 Fine-tuning Net \cite{Few_shot_transfer} & $89.70\pm 5.48\%$ & $90.58\pm 3.87\%$ & $93.18\pm 3.49\%$ & $94.31\pm 2.62\%$ & $91.94\%$ \\
Unfrozen 4 Fine-tuning Net \cite{Few_shot_transfer} & $82.41\pm 6.01\%$ & $89.24\pm 5.65\%$ & $90.49\pm 2.59\%$ & $91.64\pm 2.05\%$ & $88.52\%$ \\
Relation Net \cite{relation_net} & $\mathbf{92.33\pm 3.56\%}$ & $93.51\pm 2.40\%$ & $95.48\pm 1.40\%$ & $96.18\pm 0.89\%$ & $94.38\%$ \\
\midrule
\textbf{MAML} (learnable $\mathbf{lr}$) & $90.55 \pm 0.45\%$ & $\mathbf{94.61 \pm 0.25\%}$ & $\mathbf{96.32\pm 0.47\%}$ & $97.17 \pm 0.10\%$ & $\mathbf{94.65\%}$  \\
\bottomrule
\end{tabular}
\label{tab:PU_comparison}
\end{table*}

This study also goes beyond identifying artificially-induced bearing defects by further exploring the generalization capability of MAML in predicting real bearing failures, which are formulated as Artificial-to-Real tasks. The objective is to use a combination of artificially-damaged bearings and healthy bearings to identify those with real damages. Due to intrinsic differences between data collected from these two scenarios, standard supervised learning methods can only achieve accuracies lower than 75\% \cite{Paderborn_paper}. Additionally, these differences will also cause the transfer learning-based method to experience an obvious decrease in the average accuracy for Artificial-to-Real tasks when compared with generalizing to other artificial tasks \cite{capsule}.

The selected representative classes from the Paderborn dataset \cite{Paderborn} are listed in TABLE \ref{tab:Paderborn}, and the order of which will be strictly followed while performing different $N$-way case studies. We can conduct a maximum of $6$-way meta-testing by adapting to all of the 6 real bearing faults (classes 8 to 13). The source of meta-training data will only be the healthy (class 1) and the artificially-damaged bearings (classes 2 to 7). Both 1-shot and 5-shot tests are conducted using MAML with a fixed scalar $lr$ and with a learnable $\mathbf{lr}$ vector.

The results of identifying real bearing defects are presented in TABLE \ref{table:real}, which demonstrates satisfactory $5$-shot results with over 97\% accuracy when dealing with 3 or fewer new realistic bearing failures using MAML with learnable $\mathbf{lr}$. Fig. \ref{fig:artificial-to-real} illustrates a comparison study between TABLE \ref{table:real} and the similar $N$-way $5$-shot results obtained on the artificially-induced bearing faults in TABLE V, and it can be observed that MAML's adaptation capability to real bearing failures delivers similar, if not better results than adapting to artificially-induced faults. This robust generalization capability, though much desired, can be difficult to accomplish using other methods such as \cite{capsule} that do not involve leveraging past experience and data to learn new tasks more quickly.
%
%
%

To further validate the generalization capability of MAML in Artificial-to-Real tasks, we'll compare it with some state-of-the-art few-shot or transfer learning algorithms, such as the Unfrozen Fine-tuning Net \cite{Few_shot_transfer}, the Feature Transfer Net \cite{feature_transfer}, and the Meta Relation Net \cite{relation_net}. To make a fair comparison, we used the same Paderborn University dataset, the same data pre-processing methods (fast Fourier transform, etc), and the same source domain and target domain fault categories as \cite{Few_shot_transfer}, which is presented in TABLE \ref{tab:Paderborn_real}. Four case studies with 1-shot, 3-shot, 5-shot, and 10-shot settings are conducted.

The fault classification accuracy and standard deviation obtained using the Paderborn dataset is shown in TABLE \ref{tab:PU_comparison}. The results for the 6 few-shot transfer learning benchmark studies are obtained by running the open source code provided in \cite{Few_shot_transfer}. To reduce the effect of randomness in data sampling, the reported numerical results in this table are obtained by taking the average of 10 trials.

It can be observed from TABLE \ref{tab:PU_comparison} that the best-performing algorithm is the proposed MAML-based few-shot bearing fault diagnostic framework, though not by a large margin. This is because for the Feature Transfer Net and the 4 Unfrozen Fine-Tuning are Nets, all of the $N$ way $K$ shot data are trained with 50 epochs, or their convolutional layer weights are updated 50 times to reach a good domain adaptation performance. The Relation Net has been also updated using 100 epochs as described in \cite{Few_shot_transfer}. However, the initial parameters of MAML are only updated once. This further validates MAML’s intrinsic capability to quickly adapt to new tasks. In addition, the standard deviation of the proposed MAML-based bearing fault diagnostic framework is also 2 to 20 times lower than the benchmark studies, demonstrating the robustness and consistency of the proposed framework.
%




%
\section{Conclusion}
This paper proposed a MAML-based few-shot bearing fault diagnostic framework, which can be generalized to identify new fault scenarios not present in the meta-training data. The effectiveness of the proposed framework in identifying new artificial bearing faults was validated using the CWRU dataset, and the results demonstrated a distinct advantage over the benchmark study based on the Siamese Network. Specifically, this advantage can be up to 25\% when using MAML with learnable inner loop learning rates $\mathbf{lr}$ and a small meta-training set.

Since most of the real-world bearing failures are evolved naturally over time, we also applied the proposed method to the Paderborn dataset to further explore the generalization capability of MAML when adapting to real bearing failures. The results demonstrate that MAML is able to deliver comparable performance as adapting to artificial bearing faults. Additionally, MAML is also able to deliver competitive performance for Artificial-to-Real tasks when compared with several few-shot learning methods. These results offer promising prospects for identifying naturally-evolved bearing failures using data collected from laboratory tests with artificially-induced faults.

It is also worthwhile to mention that the vanilla MAML algorithm in \cite{MAML} may still suffer from a variety of issues, such as being sensitive to the inner loop hyper-parameters, having unstable training processes and limited generalization capabilities, among others. These issues have been extensively discussed in \cite{Train_MAML}, which also provides 6 different approaches to mitigate them. This work primarily focused on applying one of the approaches, namely learning per-step learning rates, to the vanilla MAML algorithm, and has achieved noticeable improvement in performance. Therefore, a desirable future work direction is to implement other optimization strategies in \cite{Train_MAML}, most notably on applying batch normalization parameters and optimizing on per-step target losses, to develop a more robust and strongly generalizable few-shot bearing fault diagnostic framework.
\balance
\bibliographystyle{IEEEtran}
\bibliography{IEEEabrv.bib,ref.bib} 

\end{document}